\documentclass[runningheads]{llncs}

\usepackage{amsthm}
\usepackage{amsmath,amssymb}
\usepackage{mathrsfs}
\usepackage{bm}
\usepackage{graphicx}
\usepackage[normalem]{ulem}
\useunder{\uline}{\ul}{}
\usepackage{hyperref}

\hypersetup{hypertex=true,colorlinks=true,linkcolor=blue,citecolor=blue,anchorcolor=blue}

\begin{document}
\title{Rethinking Class Activation Maps for Segmentation: Revealing  Semantic Information in Shallow Layers by Reducing Noise}
\titlerunning{Rethinking Class Activation Maps for Segmentation}
%
\author{Hang-Cheng Dong
\and
Yuhao Jiang\inst{1} \and
Yingyan Huang\inst{1} \and
Jingxiao Liao\inst{1} \and
Bingguo Liu\inst{1} \and
Dong Ye\inst{1} \and
Guodong Liu\inst{1}\thanks{Corresponding author.}
}
\authorrunning{H. Dong et al.}
%
\institute{Harbin Institute of Technology, Harbin 150001, China\\
\email{\{hunsen\_d,lgd\}@hit.edu.cn}\\
}
\maketitle              
\begin{abstract}
Class activation maps are widely used for explaining deep neural networks. Due to its ability to highlight regions of interest, it has evolved in recent years as a key step in weakly supervised learning. A major limitation to the performance of the class activation maps is the small spatial resolution of the feature maps in the last layer of the convolutional neural network. Therefore, we expect to generate high-resolution feature maps that result in high-quality semantic information. In this paper, we rethink the properties of semantic information in shallow feature maps. We find that the shallow feature maps still have fine-grained non-discriminative features while mixing considerable non-target noise. Furthermore, we propose a simple gradient-based denoising method to filter the noise by truncating the positive gradient. Our proposed scheme can be easily deployed in other CAM-related methods, facilitating these methods to obtain higher-quality class activation maps. We evaluate the proposed approach through a weakly-supervised semantic segmentation task, and a large number of experiments demonstrate the effectiveness of our approach.

\end{abstract}
\section{Introduction}

Deep neural networks have been criticized as a black box because it is difficult to understand their decision-making process~\cite{fan}. Lack of interpretability seriously hinders the application of deep learning in mission-critical domains. Fortunately, the visual explanation represented by class activation maps (CAMs)~\cite{CAM} has shed light on the interpretability of neural networks. The intuitive idea is that a better visual explanation should be more consistent with the semantics of the target. In practice it is also consistent with this intuition that CAM-based explanations generate heatmaps for CNN-based classification models where the highlighted regions have a higher probability of coinciding with the target object~\cite{CAM,GradCAM,layercam}.
Therefore, the CAM-based approaches can enable object localization and segmentation under the supervision of image-level annotation, which greatly facilitates the development of weakly supervised target localization (WSOD)~\cite{wold} and weakly supervised semantic segmentation (WSSS)~\cite{ECSnet,SEAM}.

CAM takes advantage of the characteristic that the feature maps of the last convolutional layer have discriminative semantics. By assigning appropriate weights to each channel of the feature maps, their linear combination can be computed to obtain a heatmap that reflects the position of the target. However, CAM has specific requirements for the structure of the model, i.e., a global average pooling layer that directly connects the classifiers. Grad-CAM \cite{GradCAM} is a generalized version of CAM, employing the average of the gradients as the weights of the feature maps. The weights generated in Grad-CAM are considered to represent the importance of corresponding channels with respect to the target category. Then, Grad-CAM++~\cite{Grad-cam++} reveals that using positive gradients can better indicate the features that have a positive impact on the prediction results. In addition, XGrad-CAM~\cite{Axiom-based} proposes two axioms to correct the gradient weights for CNNs with only ReLU activations. Lift-CAM~\cite{liftcam} and Relevance-CAM both use layer-wise relevance propagation (LRP)~\cite{LRP} to calculate the weights of different channels, and Relevance-CAM~\cite{RelevanceCAM} further finds that the problem of shattered gradients in shallow feature maps can be well alleviated. In addition to using the back-propagation gradient as a measure of importance, Score-CAM~\cite{scorecam} and Ablation-CAM~\cite{ablationcam} use the change in the model output value to measure the weight of the corresponding feature maps. Unlike the global weights mentioned above, Layer-CAM~\cite{layercam} suggests using back-propagated gradient matrices as local weights to highlight channel-wise feature maps. In summary, a large number of CAM variants focus on how to obtain a better weight assignment method.

Nevertheless, most of the methods mentioned above involve only the feature maps of the last convolutional layer, whose spatial resolution is very limited, resulting in a coarse localization and segmentation of the targets. A natural idea is whether it is possible to obtain semantic information about the target from shallow, high-resolution feature maps. As revealed by Relevace-CAM and Layer-CAM, shallow feature maps do contain higher-resolution semantic information, albeit with more noise. FullGrad~\cite{fullgrad} goes a step forward by fusing gradient features from all convolutional layers. For the sake of brevity, we refer to the operation of fusing features from different layers as the \textbf{layerization trick}.

Inspired by the layerization trick, the question we want to determine is whether such numerous CAM variants can be implemented with it. Moreover, although the layerization trick enables better visual explanation, it is difficult to apply in WSSS if the noise cannot be suppressed, even if the feature resolution becomes higher.

To verify the issues described above, we investigate the performance of the
basic Grad-CAM in shallow layers. Surprisingly, with only a gradient rectification, the performance of Grad-CAM can be improved to be comparable to or even surpassed by state-of-the-art methods. We further validate the gradient rectification and find that the method can also improve the performance of FullGrad and Layer-CAM, which is termed the \textbf{truncation trick}. Generally, our main contributions are threefold:

\begin{itemize} 
    \item[$\bullet$]  We propose LTGrad-CAM, which shows that global weights can still exploit shallow semantic information and are quite easy to deploy to any off-the-shelf CNN model.
    \item[$\bullet$] We propose gradient truncation methods that can enable gradient-weighted CAM-based methods to benefit from layerization tricks and reduce the disturbance of noise on weakly-supervised tasks.
    \item[$\bullet$] We summarize for the first time the layerization trick and the truncation trick. Moreover, we propose that these two techniques can be used as plug-ins for many gradient-weight CAM-based methods, which can provide fine-grained visual explanations and substantially improve the performance of weakly supervised tasks.
\end{itemize}

\section{Methodology}

In this section, to illustrate explicitly, we first review 3 directly related methods, including Grad-CAM, Layer-CAM, and FullGrad. Then we introduce our plug-and-play methods, layerization trick, and gradient truncation trick.

\subsection{Revisit Grad-CAM, Layer-CAM, and FullGrad}

Mathematically, consider a classifier $f$ with $L$ convolutional layers, whose parameters are $\bm{\theta}$. For a given input image $\bm{I}$ with category $c$, the prediction $y^c$ before the softmax can be obtained by
\begin{equation}
    y^c = f^c(\bm{I};\bm{\theta}). 
\end{equation}
Let $\bm{A}^{l}\in \mathbb{R}^{W_l \times H_l \times C_l}$ denote the feature map of the $k$-th channel generated by the $l$-th layer in the CNN, $l\in {1, 2, ..., L}$, where $W_l$ and $H_l$ are the width and height of $l$-th feature map respectively, and $C_l$ is the number of the channels in $l$-th convolutional layer. The gradient of output score $y^c$ with respect to the activation $\bm{A}^{kl}$ at location $(i,j)$ is $ g_{ij}^{ckl} = \frac{\partial y^c}{\partial A_{ij}^{kl}} $. 

\subsubsection{Grad-CAM.}When inputting an image $I$ with category $c$, Grad-CAM acquires the channel-wise weight $w_k^c$ by 
\begin{equation}
    w_{kl}^c = \frac{1}{Z_l} \sum_i \sum_j \frac{\partial y^c}{\partial A_{ij}^{kl}},
    \label{eq2}
\end{equation}
where $Z_l = W_l \times H_l$. Then, the saliency map of specific layer $l$ can be formed by a linear combination of the feature maps as

\begin{equation}
    M_{\text{Grad-CAM}}^{cl} = \text{ReLU} (\sum_k w_{kl}^c \cdot \bm{A}^{kl}),
    \label{eq3}
\end{equation}
where a ReLU function is applied to remove the negative responses from the saliency map. 

\subsubsection{Layer-CAM.}Unlike Grad-CAM, Layer-CAM calculates the local importance instead of the average of the gradient matrix corresponding to a certain feature map. Formally, the element of the weight with spatial location $(i,j)$ in the $k$-th feature map and $l$-th layer can be written as
\begin{equation}
    w_{ij}^{ckl} = \text{ReLU} (\frac{\partial y^c}{\partial A_{ij}^{kl}}).
    \label{4}
\end{equation}
Finally, Layer-CAM obtains the saliency map by
\begin{equation}
    M^{cl}_{\text{Layer-CAM}} = \text{ReLU}(\bm{w}_{kl}^{c}\otimes \bm{A}^{kl}), 
\end{equation}
where $\otimes$ represents the Hadamard product. 

\subsubsection{FullGrad.}Layer-CAM is designed to better extract the semantic features at shallow layers, whereas FullGrad analyses the piecewise linear characteristics of ReLU CNNs and suggests using the gradient features of biases in each layer. It is worth noting that only gradient information and origin inputs are in consideration for FullGrad, which is formulated as
\begin{equation}
    M^{c}_{\text{FullGrad}} = \frac{\partial y^c}{\partial \bm{x}} \otimes \bm{x} + \sum_{l=1}^{L}\sum_{k=1}^{C_l} \Psi(\frac{\partial y^c}{\partial \bm{b}^{kl}} \otimes \bm{b}^{kl}),
\end{equation}
where $\bm{x}$ is an input image, $\bm{b^{kl}}$ is the bias term of $k$-th neuron in $l$-th layer, and $\Psi(\cdot)$ is a post-processing function. Generally, $\Psi(\cdot)$ is used to align saliency maps, which is formulated as follows:
\begin{equation}
    \Psi(\bm{x}) = \text{bilinearUpscale}(\text{normalize}(\vert \bm{x} \vert)),
\end{equation}
where bilinearUpscale is a typical image interpolation operation that resizes its input image to target size, and $\text{normalize}(x_{ij}) = \frac{x_{ij}-\min\bm{x}}{\max\bm{x} - \min\bm{x}}$.

\subsection{Semantic Information from Shallow Layers}

\subsubsection{layerization Trick.} 

\begin{figure}
\includegraphics[width=\textwidth]{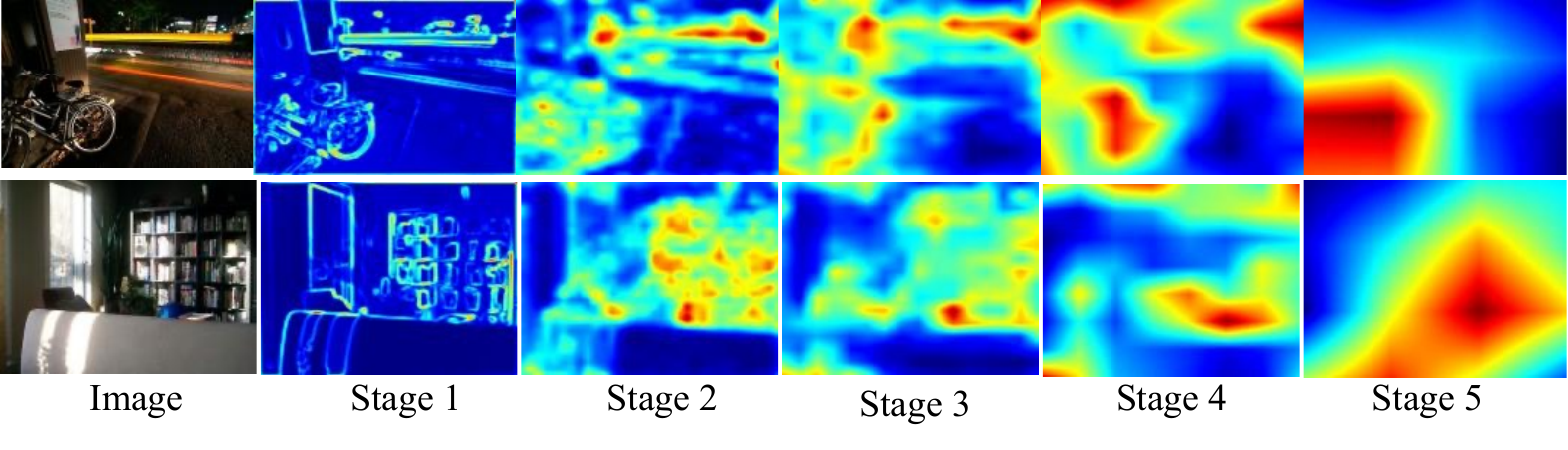}
\caption{Heatmaps generated by Grad-CAM for 5 different stages in the VGG16 model. Images are randomly selected from the PASCAL VOC 2012 dataset.} \label{fig2}
\end{figure}

Most CAM series methods obtain discriminative features from the output of the last convolutional layer. Inspired by GradFull and Layer-CAM, we visualize feature maps of the different layers from a VGG16~\cite{VGG16} network trained on the PASCAL VOC dataset~\cite{voc}. As shown in Fig.~\ref{fig2}, we generate heatmaps for the output of the convolutional layers at different stages, using the Grad-CAM method. In contrast to the view in \cite{layercam}, we observe that the high-resolution semantic features exist still in the shallow feature maps, while containing some noise. Naturally, we introduce the technique of fusing shallow feature maps to improve the resolution of CAM-based methods as follows:

\begin{equation}
    M_{\text{CAM}}^{c} = \sum_{l} \psi (M_{\text{CAM}}^{cl}),\label{eq8}
\end{equation}
where $\psi(\cdot) $ is the bilinear interpolation operation. It is worth to note that we find that the discriminability of the fused features may change, which can be solved by filtering, but for the sake of fairness we use Eq.\ref{eq8} directly to generate the heatmap in the experiments.

\subsubsection{Truncation Trick.} Fusing the high-level features, despite improving the resolution of the features, simultaneously introduces considerable noise. Therefore, we present the gradient truncation method to obtain more high-quality heatmaps. Specifically, we assume that the positive gradient reflects the features that play a positive role in classification. Moreover, the larger the gradient, the more important the features are, and the less noise there is. Formally, for all CAM series methods that rely on gradients to measure their weights, we can truncate the gradient as

\begin{equation}
    M_{\text{CAM}}^{c} = M_{\text{CAM}}^{c}(\bm{x},\mathbb{I} (\frac{\partial y^c}{\partial Z} \geq \delta) \otimes \frac{\partial y^c}{\partial Z} ),
    \label{eq9}
\end{equation}
where $\mathbb{I}(\cdot)$ is the indicator function, $Z$ is a certain target feature map, and $\delta$ is a hyperparameter that is set to the $\delta$-th percentile of the positive values in each feature map. Eventually, we obtain the heatmap by

\begin{equation}
    S_{\text{CAM}}^{c} = \sum_l \psi (M_{\text{CAM}}^{cl}(\bm{x},\mathbb{I} (\frac{\partial y^{cl}}{\partial Z} > \delta))\otimes \frac{\partial y^{cl}}{\partial Z} ).
    \label{eq10}
\end{equation}

\subsection{LTGrad-CAM}

Based on the above description, our method can be used as a plug-in for any gradient-weighted CAM-based method. In particular, combining Eqs.\ref{eq2}, \ref{eq3} and \ref{eq10}, we can directly upgrade Grad-CAM as

\begin{equation}
        S_{\text{LTGrad-CAM}}^{c} = \sum_l \psi ( \sum_k(( \frac{1}{Z_l} \sum_i \sum_j \frac{\partial y^c}{\partial A_{ij}^{kl}} \otimes
        \mathbb{I} (\frac{\partial y^{c}}{\partial A_{ij}^{kl}} > \delta)) )\cdot \bm{A}^{kl})).
\end{equation}
Similarly, Layer-CAM and FullGrad can be updated to LTlayer-CAM and LTFullGrad, respectively. 

We would like to emphasize that although Layer-CAM has proposed the idea of fusing different layers, Layer-CAM believes that the layerization trick is not suitable for other CAM methods. Our study corrects this point, and reveals that layerization tricks are potentially useful in different scenarios.

\section{Experiments}

\subsection{Experimental Setup}

\subsubsection{Datasets.}We conduct weakly-supervised semantic segmentation experiments of the proposed method on two datasets. The first is a common object detection dataset, the PASCAL VOC 2012 dataset, and this dataset contains 20 semantic categories and backgrounds. The original images are divided into 1464 training images, 1449 validation images, and 1456 test images. The second dataset is the industrial surface defect detection dataset KSDD2~\cite{ksdd2}. This dataset consists of over 3000 images with approximately 230 pixels in width and 630 pixels in height. The dataset is split into a training set with 2085 defect-free samples and 246 defective samples, and a test set with 894 defect-free samples and 110 defective samples.

\subsubsection{Implementation Details.} We train the classification models on each dataset with image-level labels only, which adopts the VGG16 model pre-trained on ImageNet~\cite{imagenet}. We employ the Otsu method to process the class activation maps to produce the final segmentation results. All experiments are executed on a PC with an NVIDIA RTX 2080Ti. We prefix the method with "LT" to indicate that the method uses the layerization trick and the truncation trick, $l$ is the number of layers fused, and $\delta$ is the percentile of the truncated positive gradient. To be fair, we only get the output of all convolution layers in the FullGrad-related method, while the other methods only get the output of the stages by default.

\subsubsection{Evaluation Metrics.} Weakly-supervised semantic segmentation requires higher pixel accuracy and hence a higher requirement for the fine granularity of the class activation maps. We adopt the mean intersection-over-union (mIoU) to evaluate the segmentation results and use pixel recall, accuracy, and Mirco-F1 metrics applicable to multi-classification tasks for a comprehensive evaluation.

\subsection{Effectiveness of the layerization Trick}

We test on the PASCAL VOC dataset whether fusing different layers is effective for the Grad-CAM method. As shown in Tab.\ref{tab1}, S1-S5 denote the 5-stage outputs of the VGG16 network. What can be identified is that fusing multiple layers is more effective than shallow features, but weaker than deep features S5. It indicates that the straightforward use of the layerization trick may be disadvantageous.

\begin{table}
\centering
\caption{Comparison of segmentation performance of class activation maps generated by different layers of VGG16 using Grad-CAM.}\label{tab1}
\begin{tabular}{c|c|c|c|c|c|c|c}
\hline
block & S1    & S2    & S3    & S4    & S5             & S3+S4+S5 & S1+S2+S3+S4+S5 \\ \hline
mIoU  & 10.13 & 10.87 & 11.68 & 14.03 & \textbf{25.31} & 22.84    & 21.57          \\ \hline
\end{tabular}
\end{table}

\subsection{Effectiveness of Truncation Trick}

Gradient truncation is tested on the PASCAL VOC dataset for Grad-CAM and FullGrad. As shown in Tab.\ref{tab2}, gradient truncation (LTGrad-CAM($l=1,\delta=0$)) for Grad-CAM generated from S5 feature maps results in a 0.92\% improvement in mIoU. Using positive gradient truncation for FullGrad, namely LTFullGrad($\delta=0$), the mIoU remains almost the same, with a slight decrease of 0.09\%. This indicates that direct employment of positive gradient truncation is more likely to have a beneficial effect. We will discuss the effect of the hyperparameter $\delta$ of the truncation trick in section~\ref{section:3.5}.

\begin{table}[]
\centering
\caption{Comparison of different methods and their performance after gradient truncation.}
\label{tab2}
\begin{tabular}{c|c|c|c|clll}
\cline{1-5}
method              & mIoU           & Precision      & Recall        & Micro-F1       &  &  &  \\ \cline{1-5}
Grad-CAM++          & 20.32          & \textbf{47.75} & 27.43         & 33.06          &  &  &  \\
Grad-CAM            & 25.31          & 40.93          & 42.02         & 39.5           &  &  &  \\
LTGrad-CAM($l=1,\delta=0$) & \textbf{26.23} & 42.46          & \textbf{44.1} & \textbf{40.59} &  &  &  \\ \cline{1-5}
FullGrad            & \textbf{28.2}  & \textbf{48.72} & 42.52         & \textbf{42.84} &  &  &  \\ 
LTFullGrad($\delta=0$)     & 28.11          & 48.01          & \textbf{42.9} & 42.77          &  &  &  \\ \cline{1-5}
\end{tabular}
\end{table}

\subsection{Comparison with State-of-the-Art Methods}

\begin{figure}
\includegraphics[width=\textwidth]{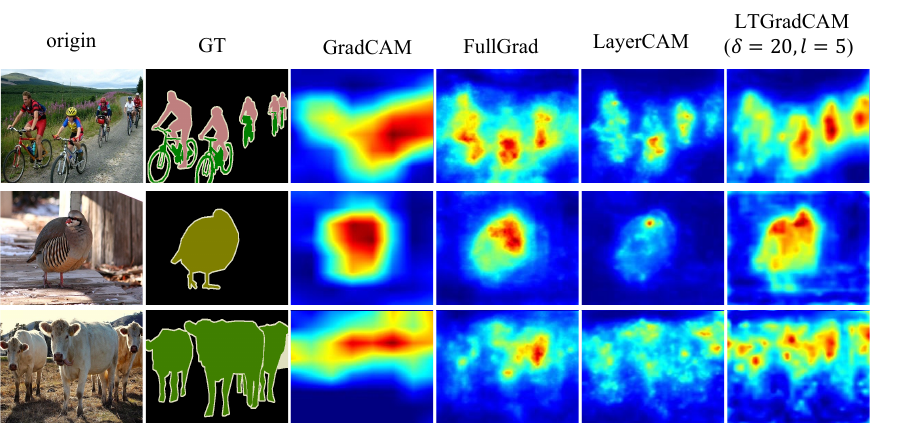}
\caption{Visualization of class activation maps generated by different methods.} \label{fig4}
\end{figure}

We report the evaluation results of weakly-supervised semantic segmentation on the object detection dataset and the defect detection dataset, respectively. Tab.\ref{tab3} demonstrates the segmentation performance on the PASCAL VOC dataset. We set the hyperparameters a to 0 and 20, and the selection principle can be based on the results of the training or validation sets, or refer to the analysis in Sec.~\ref{section:3.5} to select a small non-zero value. As shown in Fig.\ref{fig4}, we show the visualization results of LTGradCAM with other methods. It can be found that our method does capture more semantic information.

Our proposed method LTGrad-CAM, as a plug-in, significantly improves the performance of Grad-CAM , which exceeds that of FullGrad, one of the sota methods, to a comparable level with Layer-CAM. It is also worth pointing out that Layer-CAM ignores the fact that shallow feature maps can also work well for other CAM-based methods.

\begin{table}[]
\centering
\caption{Performance comparison on weakly-supervised semantic segmentation experiments on the PASCAL VOC dataset. The results in bold indicate the strongest performance, with underlined ones indicating the second highest performance.}
\label{tab3}
\begin{tabular}{c|c|c|c|clll}
\cline{1-5}
method                                    & mIoU           & Precision      & Recall         & Micro-F1       &  &  &  \\ \cline{1-5}
Grad-CAM++                                & 20.32          & 47.75          & 27.43          & 33.06          &  &  &  \\
Grad-CAM                                  & 25.31          & 40.93          & 42.02          & 39.5           &  &  &  \\
FullGrad                                  & 28.2           & {\ul 48.72}    & 42.52          & 42.84          &  &  &  \\
Layer-CAM($l=1,\delta=0$)                        & 26.14 & 43.41 & 43.0           & 40.5 &  &  &  \\ 
Layer-CAM($l=5,\delta=0$)                        & \textbf{30.83} & \textbf{52.07} & 45.3           & \textbf{45.95} &  &  &  \\ \cline{1-5}
\multicolumn{1}{l|}{LTGrad-CAM($l=5,\delta=0$)}  & 26.87          & 40.86          & {\ul 48.56}    & 41.24          &  &  &  \\
\multicolumn{1}{l|}{LTGrad-CAM($l=5,\delta=20$)} & {\ul 28.34}    & 43.39          & \textbf{48.78} & {\ul 43.06}    &  &  &  \\ \cline{1-5}
\end{tabular}
\end{table}

\begin{figure}
\includegraphics[width=\textwidth]{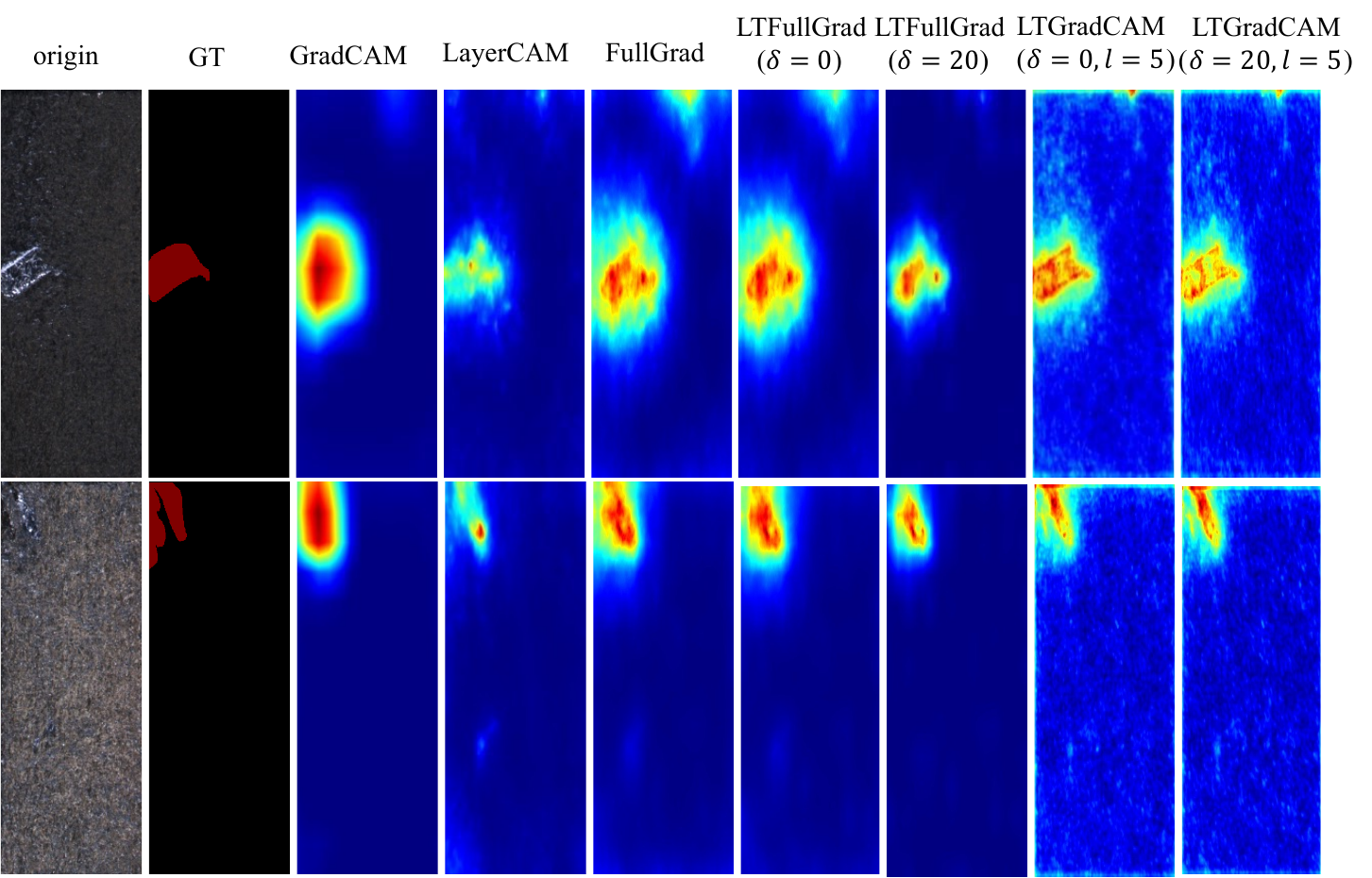}
\caption{Visualization of class activation maps generated by different methods on the KSDD2 dataset.} \label{fig5}
\end{figure}

As shown in Tab.\ref{tab4}, the superiority of LTGrad-CAM is more evident in KSDD2 dataset. In the most fundamental configuration, an outperforms the mIoU score of the state-of-the-art Layer-CAM by 4.09\%, and this gap further expands to 6.45\% when $\delta=20$. In terms of the micro-F1 metric, our method improves by 7.4\% over Layer-CAM($l=5,\delta=0$). As shown in Fig.\ref{fig5}, we show in detail the heatmap produced by different methods on the KSDD2 dataset.

\begin{table}[]
\centering
\caption{Performance comparison on weakly-supervised semantic segmentation experiments on the KSDD2 dataset. The results in bold indicate the strongest performance, with underlined ones indicating the second highest performance.}
\label{tab4}
\begin{tabular}{c|c|c|c|clll}
\cline{1-5}
method                                    & mIoU                         & Precision                    & Recall                                & Micro-F1                     &                                             &  &  \\ \cline{1-5}
Grad-CAM++                                & 25.87                        & 34.75                        & 50.31                                 & 41.11                        & \multicolumn{1}{c}{}                        &  &  \\
Grad-CAM                                  & 25.73                        & 27.79                        & {\ul 77.63}                           & 40.93                        &                                             &  &  \\
FullGrad                                  & { 22.14} & { 23.41} & { \textbf{80.25}} & { 36.25} & \multicolumn{1}{c}{{ }} &  &  \\
Layer-CAM($l=1,\delta=0$)                       & 20.50 & 21.67 & 79.14           & 34.03 &  &  &  \\ 
Layer-CAM($l=5,\delta=0$)                        & 28.92              & 31.8             & 76.12                                 & 44.86              &                                            &  &  \\ \cline{1-5}
\multicolumn{1}{l|}{LTGrad-CAM($l=5,\delta=0$)}  & {\ul 33.01}                  & {\ul 37.07}                  & 75.06                                 & {\ul 49.63}                  &                                             &  &  \\
\multicolumn{1}{l|}{LTGrad-CAM($l=5,\delta=20$)} & {\ul \textbf{35.37}}         & \textbf{40.88}               & 72.42                     & \textbf{52.26}               &                                             &  &  \\ \cline{1-5}
\end{tabular}
\end{table}

\subsection{Influence of the Truncation Hyperparameters $\delta$} \label{section:3.5}

\begin{figure}
\includegraphics[width=0.85\textwidth]{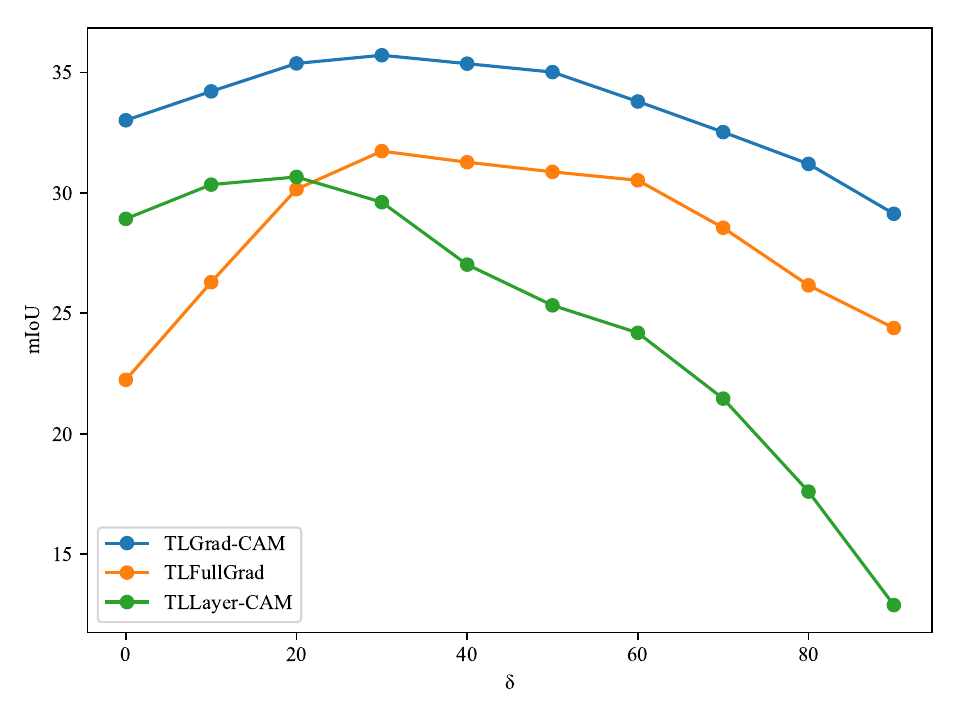}
\caption{Variation of semantic segmentation performance of three methods with different truncation values $\delta$ on the testing set. The blue line indicates Grad-CAM, the orange line indicates FullGrad, and the green line indicates Layer-CAM.} \label{fig3}
\end{figure}

\begin{figure}
\includegraphics[width=0.85\textwidth]{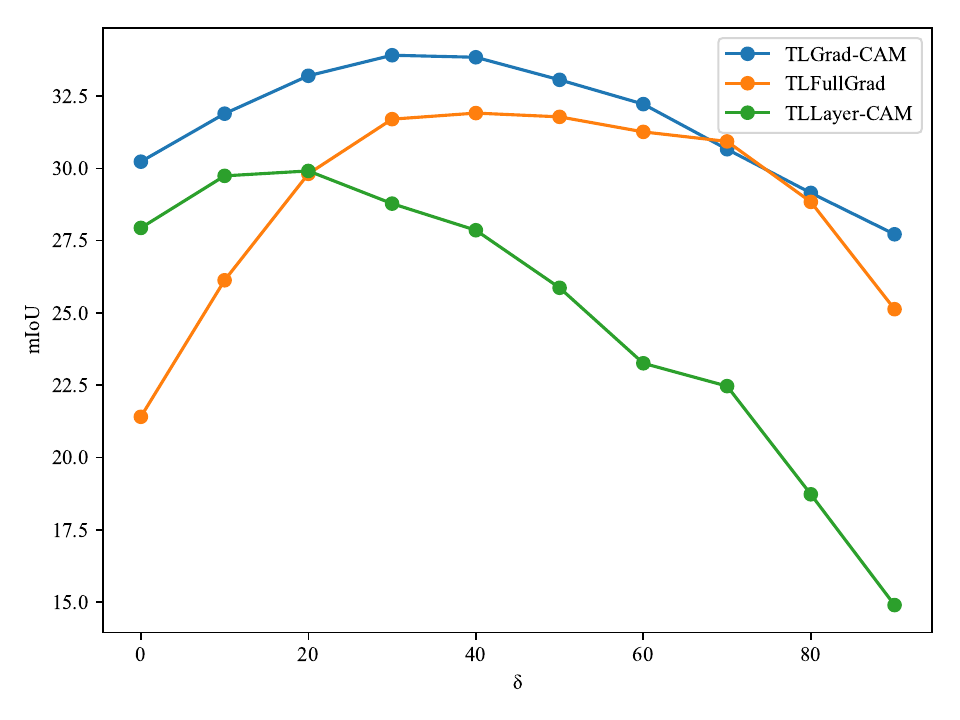}
\caption{Variation of semantic segmentation performance of three methods with different truncation values $\delta$ on the training set. The blue line indicates Grad-CAM, the orange line indicates FullGrad, and the green line indicates Layer-CAM.} \label{fig6}
\end{figure}

The hyperparameter $\delta$ is critical to the proposed method. We test different truncation percentages on the KSDD2 dataset, sampled at 10 intervals from 0-90, and plot the variation in the mIoU scores on the training and testing sets. We chose three typical methods for truncation, namely Grad-CAM, LayerCAM, and FullGrad. As shown in Fig.\ref{fig3}, all three curves show an increasing trend followed by a decreasing trend. An obvious explanation is that as the truncation quantile increases, the noise that contributes little to the predicted value is filtered out first. Subsequently, the target features also start to decrease. Another positive sign is that the performance after gradient truncation is generally improved when the quantile lies between 0 and 40. As shown in Fig.\ref{fig6}, the trend on the training set is very similar to that on the testing set, i.e., a suitable gradient truncation hyperparameter can be found on the training set. This indicates that our proposed method, which can be used as a plug-and-play module, may help other related CAM methods benefit from this concise operation. Finally, our proposed LTGrad-CAM outperforms existing methods by a large margin in defect inspection tasks, which also suggests that earlier proposed CAM methods may have additional potential.

\section{Conclusion}
In this work, we have revisited the key factors affecting the quality of class activation maps and proposed to filter out the noise in the shallow feature maps by a gradient truncation technique, which makes it possible to fuse the semantic information of different layers. Experiments have demonstrated that the proposed method can be well combined with various gradient-weighted CAM-based related methods. In the future, we will investigate how to further enhance the resolution of CAM methods to further improve the performance of weakly-supervised semantic segmentation.

%
%

%
%
%
\bibliographystyle{splncs04}
\bibliography{mybib}
%




\end{document}